\pdfoutput=1

\documentclass[11pt]{article}

\usepackage[final]{acl}

\usepackage{geometry}
\usepackage{array}
\usepackage{times}
\usepackage{latexsym}
\usepackage{tabularx}
\usepackage{pgfplotstable}
\usepackage{booktabs} 
\usepackage{multirow} 
\usepackage{array}
\usepackage{graphicx}
\usepackage{algorithm}
\usepackage{algorithmic}
\usepackage{amsmath}
\usepackage[english]{babel}
\usepackage{amsthm}
\usepackage{amsfonts}
\usepackage{svg}
\usepackage{colortbl}
\usepackage[breakable,skins]{tcolorbox}

\PassOptionsToPackage{dvipsnames,table,usenames}{xcolor}

\usepackage[T1]{fontenc}

\usepackage[utf8]{inputenc}

\usepackage{microtype}

\usepackage{inconsolata}

\usepackage{graphicx}

\title{Reasoning Aware Self-Consistency: Leveraging Reasoning Paths for Efficient LLM Sampling }

\author{
Guangya Wan$^{1}$\thanks{These authors contributed equally to this work.},
Yuqi Wu$^{2}$\footnotemark[1],
Jie Chen$^{2,\dagger}$,
Sheng Li$^{1,\dagger}$\\
$^1$School of Data Science, University of Virginia\\
$^2$Department of Electrical and Computer Engineering, University of Alberta\\
\{wxr9et,shengli\}@virginia.edu,
\{yuqi14,jc65\}@ualberta.ca\\
\thanks{$\dagger$Corresponding authors.}
}

\begin{document}
\maketitle
\begin{abstract}
Self-Consistency mitigates hallucinations in Large Language Models (LLMs) by sampling multiple reasoning paths, but it lacks a systematic approach to determine the optimal number of samples or select the most faithful rationale. To address this limitation, we introduce Reasoning-Aware Self-Consistency (RASC), a novel framework that enhances sampling efficiency and reasoning faithfulness by dynamically evaluating both outputs and rationales. RASC assesses the quality of reasoning and the consistency of answers for each generated sample, using these assessments to guide early stopping decisions and rationale selection.The framework employs criteria-based stopping and weighted majority voting, enabling more informed choices on when to halt sampling and which rationale to select. Our comprehensive experiments across diverse question-answering datasets demonstrate that RASC outperforms existing methods, reducing sample usage by approximately 70\% while maintaining accuracy. Moreover, RASC facilitates the selection of high-fidelity rationales, thereby improving the faithfulness of LLM outputs. Our approach effectively addresses the efficiency-accuracy trade-off in LLM reasoning tasks, offering a new perspective for more nuanced, faithful, and effective utilization of LLMs in resource-constrained environments.~\footnote{Code available at: \url{https://github.com/wan19990901/RASC/tree/Submission_Code}}
\end{abstract}

\section{Introduction}
\begin{figure*}[t]
  \includegraphics[width=\linewidth]{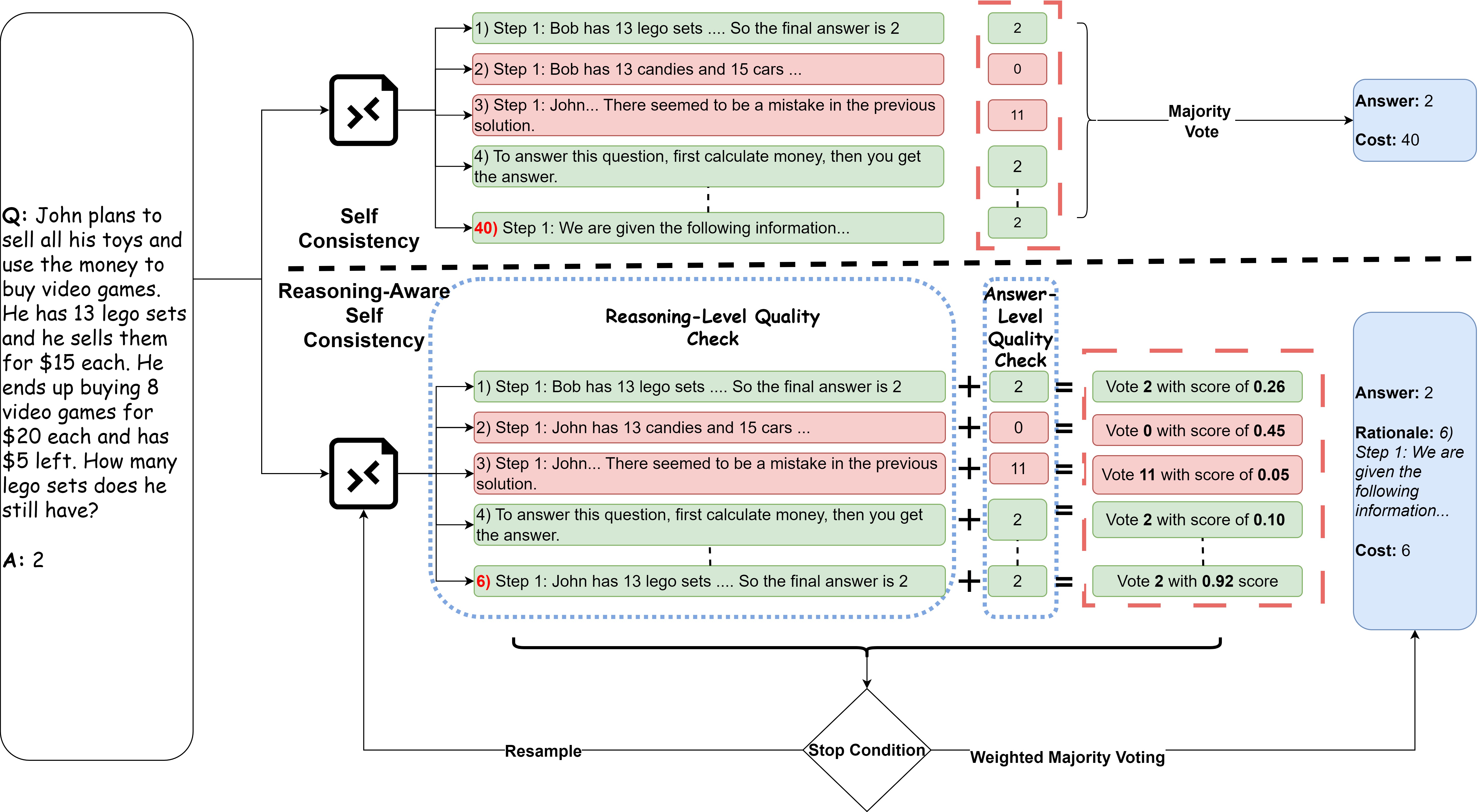} 
 \caption{Comparison between Self Consistency and Reasoning-Aware Self Consistency (RASC) methods for a reasoning problem. The top half illustrates the standard Self Consistency approach, generating multiple reasoning paths and using majority voting for the final answer. The bottom half demonstrates RASC's improvements: it assigns scores based on the qualities of both answers and reasoning paths, effectively handling incorrect or irrelevant responses. RASC's resampling and stop condition mechanisms optimize sampling, enhancing efficiency and maintaining accuracy. Moreover, RASC enables the selection of the most faithful rationale among generated samples, improving the quality and reliability of the reasoning process in complex tasks}
  \label{fig:graph_abs}
\end{figure*}

Large Language Models (LLMs) have demonstrated remarkable complex reasoning capabilities, particularly when employing chain-of-thought (CoT) prompting \citep{wei2022chain} which allows LLMs to generate reasoning paths (RPs) before answers. Building upon this, \citet{wang2022self} introduced Self-Consistency (SC), a decoding strategy that significantly enhances reasoning performance by sampling multiple RPs and marginalizing over these samples. This approach has substantially improved LLM's performance across various reasoning domains \citep{tan2023can,ahn2024large}.
However, SC's effectiveness is tied to the number of reasoning paths sampled, creating a critical trade-off between performance and computational cost. While the original paper show that sampling more paths (e.g., 40) generally leads to better performance, there's no systematic method to determine the optimal number of samples for any given task. This ambiguity often results in over-sampling, leading to excessive computational demands \citep{li2024escape}. Thus, we need more efficient methods that can maintain SC's reasoning benefits while reducing its computational burden.

Recent research has introduced early-stopping mechanisms to address the computational costs of Self-Consistency. Adaptive Consistency (AC) \citep{aggarwal2023let} employs incremental sampling, terminating when a clear majority emerges based on a predefined probability distribution, while Early Stopping Self-Consistency (ESC) \citep{li2024escape} segments the preset sample size into windows and stops sampling when answers within a window achieve uniformity. Although these methods effectively reduce computational demands, they fail to address two critical limitations in the original self-consistency approach: (1) they treat all samples equally for sampling decisions, basing the stopping criteria solely on the consistency of final answers, with this uniform treatment extending to the use of majority voting; (2) these approaches lack of a formal mechanism to distinguish and prioritize the most faithful and high-quality rationale among the selected samples, which can be critical in domains that require precise explanations such as healthcare and science \cite{wang2024prompt, lu2022learn}. 

To address these limitations while preserving the benefits of self-consistency, we propose Reasoning-Aware Self-Consistency (RASC), a framework that enhances the original SC approach. As illustrated in Fig.~\ref{fig:graph_abs}, our method assesses both the rationales across samples and the consistency of final answers, which enables more nuanced and efficient sampling. RASC is motivated by the principle that higher-quality rationale often leads to better LLM reasoning \cite{zelikman2022star}. By leveraging this insight, RASC prioritizes robust reasoning in decision-making and enables early stopping when a consistent pattern of high-quality RP and consistency of answers is observed. This approach not only improves  efficiency but also allows RASC to identify the most faithful and high-quality chains of thought among selected samples. Our approach significantly outperforms existing methods across 10 datasets spanning Commonsense, Mathematical, and Symbolic reasoning, reducing sample usage by 60\%-80\% while maintaining or improving accuracy. Furthermore, experiments demonstrate RASC's ability to select high-fidelity chain-of-thought reasoning, as well as its robustness across various datasets, prompting strategies, and multiple base LLM models, highlighting its adaptability to different efficiency-accuracy trade-offs and application needs. This combination of efficient sampling, high-fidelity reasoning selection, and broad applicability positions RASC as a versatile and powerful tool for enhancing the reasoning capabilities of large language models. In summary, our key contributions are as follows:

\begin{itemize}
    \item \textbf{Analysis of Sampling-Based LLM Reasoning Techniques}: We present the first systematic evaluation of current SC-based sampling methods for LLM reasoning, offering new insights into their limitations, including the disregard for reasoning quality, uniform sample weighting, and the inability to identify the most faithful rationale. 
    
    \item \textbf{Reasoning-Aware Self-Consistency (RASC) Framework}: We propose a novel framework that accounts for both rationale quality and answer consistency to dynamically select the number of samples and the best rationale for sampling-based LLM reasoning .
    
    \item \textbf{Robust Evaluation}: We showed RASC outperforms existing techniques in efficiency and explanation quality across various settings.
\end{itemize}

\section{Related Work}

\subsection{Self-Consistency and Sampling Methods}
Self-Consistency (SC) \citep{wang2022self} is a powerful decoding strategy that mitigates LLM hallucinations by sampling multiple reasoning paths and deriving final answers through majority voting. Recent studies have demonstrated SC's critical importance and versatility across different domains: \citet{huang2024large} showed that SC outperforms alternative approaches like multi-agent debate for reasoning tasks, while \citet{chen2024universal} and \citet{wang2024integrate} successfully extended SC to open-ended generation tasks including reasoning domains and text summarization. However, despite its effectiveness, SC incurs significant computational costs at inference time due to its requirement for multiple sampling iterations. To address this limitation, researchers have proposed several adaptive approaches: Adaptive Consistency (AC) \citep{aggarwal2023let} employs incremental sampling with a Dirichlet-based stopping criterion, while Early-Stopping Self-Consistency (ESC) \citep{li2024escape} segments sampling into sequential windows and stops when answers achieve uniformity. Although these methods effectively reduce computational demands, they primarily focus on checking the consistency of answers but overlook sample quality variations and the reasoning process's importance. Our work extends these approaches by addressing such limitations, aiming to enhance both the computational efficiency and faithfulness of self-consistency techniques.

\subsection{Chain-of-Thought Prompting and Reasoning Process}

Chain-of-Thought (CoT) prompting \citep{wei2022chain} has significantly enhanced LLM performance through step-by-step reasoning, with variants \citep{kojima2022large,zhou2023leasttomost} further refining this approach. Recent work has sparked the idea that higher-quality intermediate reasoning paths lead to better final answers \citep{zelikman2022star,zhang2024restmcts,khan2024debatingpersuasivellmsleads}, concerns about the faithfulness of CoT-generated explanations have also emerged, particularly the risk of hallucination snowballing \citep{zhang2024hallucinations}. This has prompted a research theme on how to evaluate the intermedaite reasoning paths \citep{lightman2024lets, radhakrishnan2023questiondecompositionimprovesfaithfulness, golovneva2023roscoe} and developing methods for reasoning path correction for better reasoning \citep{wan2024cotrerailerenhancingreliability, miao2024selfcheck, weng2023large}.Building on these advancements, our work introduces RASC, which adapts CoT evaluation frameworks to the self-consistency setting to dynamically enhancing the reliability and faithfulness of SC-based reasoning.

\section{Reasoning-Aware Self-Consistency}

Self-Consistency (SC) and its variants enhance question-answering accuracy in Large Language Models (LLMs) by generating multiple answers, but often overlook reasoning processes, leading to inefficient sampling. We propose Reasoning-Aware Self-Consistency (RASC), a framework that optimizes sampling efficiency while maintaining accuracy by evaluating both reasoning paths and answers. RASC introduces a sufficiency scoring function that combines reasoning quality and answer consistency features, mapped to a continuous score via an optimized classifier. The framework employs a dynamic sampling process with a buffer of high-quality samples, stopping when a predefined capacity is reached. Finally, RASC uses weighted majority voting based on the scores to select the final answer and most faithful reasoning path. This approach integrates reasoning evaluation into the self-consistency process, offering a more faithful and computationally efficient method for complex question-answering tasks in LLMs.

\begin{table*}[h!]
    \centering
    \caption{Answer-Level and Reasoning-Level Features for RASC. These features were designed based on our preliminary analysis (Appendix \ref{app:feature}) and insights from previous literature \citep{jin2024impact, li2024dissecting, zhang2023siren,huang2023survey,hosking2023human,bang2023gptcache,tu2020approximate,golovneva2023roscoe}. They capture various aspects of reasoning quality and answer consistency. Symbols in parentheses indicate relationships between samples: *-Consistency targets exact matches, while -Relevance considers vocabulary overlap. This comprehensive set of features enables RASC to make informed decisions about sampling sufficiency. Refer to Table \ref{tab:features-example} in appendix for specific examples. }
    \label{tab:features}
    \small  
    \resizebox{\textwidth}{!}{  
    \begin{tabular}{>{\raggedright\arraybackslash}p{0.22\textwidth}  
                    >{\raggedright\arraybackslash}p{0.50\textwidth} 
                    >{\raggedright\arraybackslash}p{0.2\textwidth}} 
        \toprule
        \textbf{Feature} & \textbf{Description}& \textbf{Calculation} \\
        \midrule
        \multicolumn{2}{l}{\textbf{Answer-Level Features}} \\  
        \cmidrule(l){1-3} 
        \textbf{Local-Consistency} ($a_{t} \leftrightarrow a_{t-1}$) & Check if the current answer is the same as the prior answer. & $LC = 1$ if $a_t= a_{t-1}$, otherwise $LC = 0$\\
        \textbf{Global-Consistency} ($a_{t} \leftrightarrow a_1,...,a_{t-1}$)& Check if the current answer is within the previous answers. & $GC = 1$ if $a_t \in A$, otherwise $GC = 0$ \\
        \textbf{Parsing-Error} ($a_t$) &  Ans: error $\rightarrow$ Parsing-Error = 1; Ans: 2 $\rightarrow$ Parsing-Error = 0 &$PE = 1$ if the answer is an error, otherwise $PE = 0$\\
        \midrule
        \multicolumn{2}{l}{\textbf{Reasoning-Level Features}} \\  
        \cmidrule(l){1-3} 
        \textbf{RP-Length} & Reasoning path string length. &$\text{RP-Length} = \text{len}(RP)$\\
        \textbf{Num-of-Steps} ($r_t$) & Number of steps in reasoning paths.&$\text{Num-of-Steps} = \text{count}(\text{steps})$ \\
        \textbf{Step-Relevance} ($r_t$) & The coherence between the reasoning steps (number of overlapping words). &$SR = \frac{|A \cap B|}{|A \cup B|}$
where $A = {\text{vocab in step}_{t-1}}$, $B = {\text{vocab in step}_t}$\\
        \textbf{Question-Relevance} ($r_t \leftrightarrow Q$) & The similarity between the input question $Q$ and the reasoning path. &$QR = \frac{|A \cap B|}{|A \cup B|}$
where $A = {\text{vocab in question}}$, $B = {\text{vocab in RP}}$\\
        \textbf{Error-Admitting} ($r_t$) & The LLM acknowledges making mistakes during the response. & $EA = 1$ if ERROR else $EA=0$\\
        \textbf{Local-Relevance} ($r_t \leftrightarrow r_{t-1}$) & The similarity between the current and previous generated reasoning paths. &$LR = \frac{|A \cap B|}{|A \cup B|}$
where $A = {\text{vocab in } RP_{t-1}}$, $B = {\text{vocab in } RP_t}$\\
        \textbf{Global-Relevance} ($r_t \leftrightarrow r_1, ..., r_{t-1}$) & The similarity between the concatenation of all previous sample RPs and the current sample RP. &$GR = \frac{|A \cap B|}{|A \cup B|}$
where $A = {\text{vocab in } RP_{1-t-1}}$, $B = {\text{vocab in } RP_t}$\\
        \bottomrule
    \end{tabular}
    }
\end{table*}

\subsection{Reasoning and Answer Level Features}

We formalize our sufficiency scoring approach by defining a function $F: X \rightarrow [0,1]$, where $X$ represents the space of all possible reasoning-answer pairs. This score is a composite measure of reasoning quality and answer consistency, indicating whether it's sufficient to terminate sampling. For a given reasoning-answer pair $x = (r, a) \in X$, we approximate $F(x)$ through a composition of functions: $F(x) \approx f_{\theta^*}(\phi(x))$.
Here, $\phi: X \rightarrow \mathbb{R}^n$ is a feature extraction function mapping the reasoning-answer pair to an $n$-dimensional real-valued vector, and $f_{\theta^*}: \mathbb{R}^n \rightarrow [0,1]$ is an optimized classifier that outputs a continuous sufficiency score given the $n$-dimensional input value. To capture both reasoning quality and answer consistency, we design $\phi$ to extract two sets of features:

\noindent\textbf{Reasoning Quality Features:} These features assess the quality of the RPs generated through CoT prompting. As shown in Table~\ref{tab:features}, these include measures such as RP-Length, Num-of-Steps, Step-Relevance, and Question-Relevance. These features capture the coherence, relevance, and depth of the reasoning process. In this study, the relevance was calculated as Jaccard Similarity.

\noindent\textbf{Answer Consistency Features:} These features, including Local-Consistency and Global-Consistency as shown in Table~\ref{tab:features}, evaluate the consistency of answers in the generated samples.

The feature extraction function $\phi$ now combines both Reasoning Quality and Answer Consistency features into a single vector:
$\phi(x_i) = [\text{Local Consistency}(x_i), ..., \text{Global Relevance}(x_i)]$
\noindent\textbf{Feature Set Applicability and Extensibility:} Our feature set provides a strong foundation for evaluating reasoning quality and answer consistency. We designed $\phi$ with a modular architecture to allow future integration of additional features as our understanding evolves. Importantly, these features are intentionally lightweight, avoiding computationally intensive extractors to optimize the cost of SC while maintaining effectiveness.

\begin{figure}[t]
  \includegraphics[width=\columnwidth]{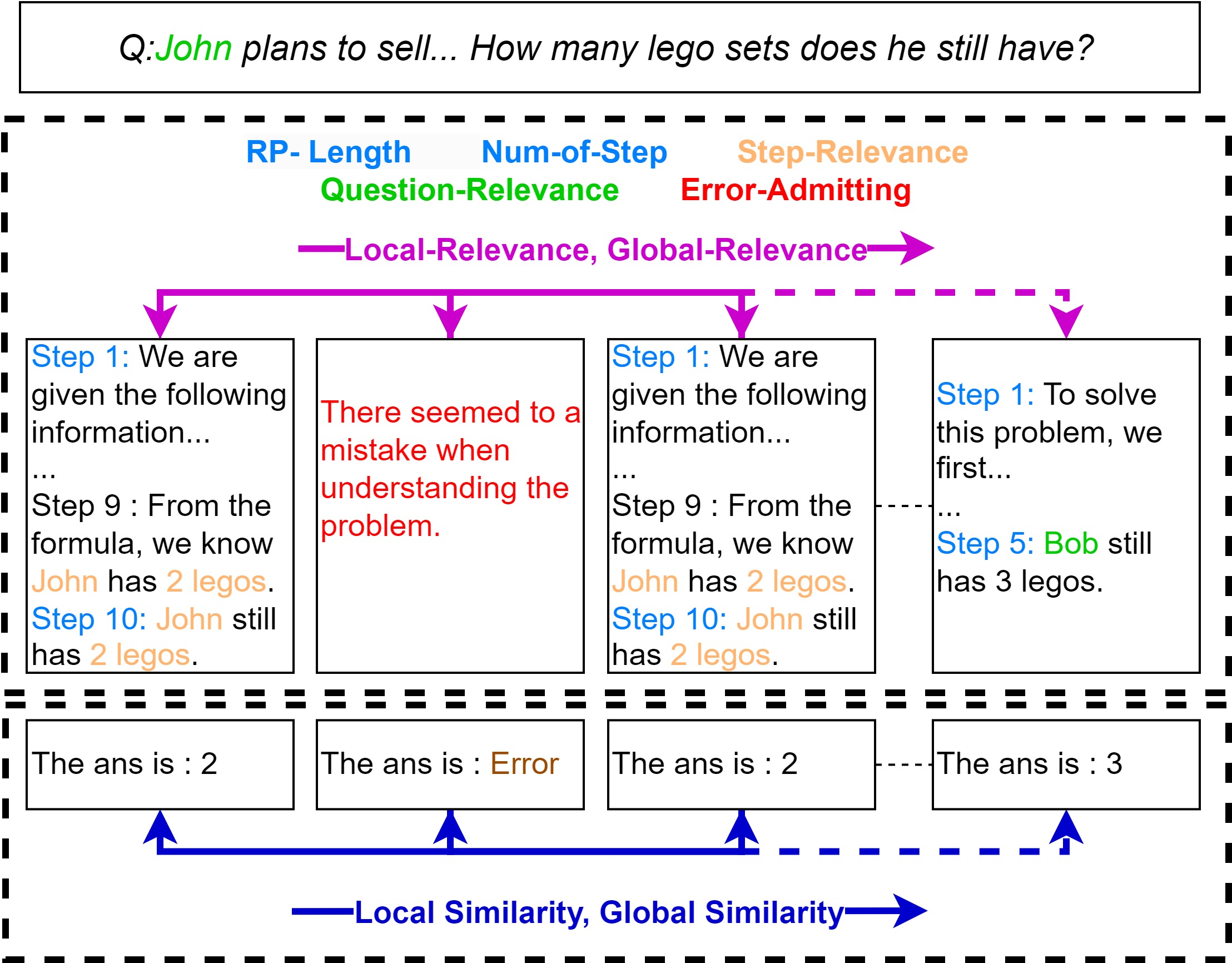}
  \caption{Reasoning-Aware quality features. Different colour corresponds to different feature visualizations. Ans: Output Answer. The upper dashed box is the reasoning-level features, and the lower dashed box is the answer-level features. The blue and purple arrows represent relative relations between different samples (consistency and relevance).}
  \label{feature}
\end{figure}

\subsection{Sampling and Decision Process}
\noindent\textbf{Sufficiency Score Computation}: We learn a scoring function $f: \mathbb{R}^n \rightarrow [0, 1]$ from a family of parameterized functions $\mathcal{F}$, where each function is denoted as $f_\theta$ for parameters $\theta$. Given a dataset $\mathcal{D} = {(x_i, y_i)}_{i=1}^{M}$, where $x_i = \phi(r_i, a_i)$ is the feature vector of the $i$-th reasoning-answer pair $(r_i, a_i)$, and $y_i \in {0, 1}$ indicates the correctness of the answer, we obtain the optimal parameters $\theta^*$ by minimizing:
\begin{equation}
\theta^* = \arg \min_{\theta} \left( \frac{1}{M} \sum_{i=1}^{M} \mathcal{L}(f_{\theta}(x_i), y_i)\right)
\end{equation}
Where $\mathcal{L}$ represents the loss function (e.g., cross-entropy).

\noindent\textbf{Stop Condition}: For a sequence of $K$ sampled reasoning-answer pairs ${x_i}_{i=1}^{K}$, we compute sufficiency scores ${SS_i = f_\theta(\phi(x_i))}$ using the optimized function $f_\theta$. We maintain a buffer $\mathcal{B}$ of high-quality pairs with sufficiency scores above a predefined threshold $T$:

\begin{equation}
\mathcal{B} = {(r_i, a_i) \mid SS_i \geq T}
\end{equation}
The sampling process terminates when $|\mathcal{B}| \geq N$, where $N$ is a predefined capacity.

\noindent\textbf{Key Trade-off Parameters}: $N$ and $T$ are crucial for balancing sample reduction and accuracy. $N$ determines the number of high-quality samples required before stopping, while $T$ sets the threshold for considering a sample as high-quality. Adjusting these parameters allows  RASC to prioritize either greater sample reduction or higher accuracy.

\noindent\textbf{Weighted Sampling for Answer and Rationale}: Upon reaching the stop condition, RASC determines the final answer and reasoning path through weighted majority voting using the sufficiency scores of pairs in $\mathcal{B}$:
\begin{align}
(\text{Ans}, \text{RP}) = \bigg(&\underset{a \in \mathcal{A}}{\arg\max} \left( \sum_{(r_i, a_i) \in \mathcal{B}, a_i = a} SS_i \right), \nonumber \\
&\underset{r_j : (r_j, \text{Ans}) \in \mathcal{B}}{\arg\max} SS_j\bigg)
\end{align}
where $\mathcal{A}$ represents the set of possible answers, $\mathcal{B}$ is the buffer of high-quality reasoning-answer pairs, and $SS_i$ is the sufficiency score for the $i$-th pair. This approach ensures that higher-quality reasoning-answer pairs have greater influence on the final decision. The answer (Ans) with the highest weighted sum of sufficiency scores is selected, and then the most faithful reasoning path (RP) is chosen as the one with the highest sufficiency score among those supporting the selected answer. For complete details, refer to Algorithm~\ref{algo_RASC} and Appendix \ref{app:theory_rasc} for theoretical justification. 

\begin{algorithm}[t]
\caption{RASC Early Stopping}\label{algo_RASC}
\begin{algorithmic}[1]
\STATE \textbf{Input:} Query $Q$, Threshold $T$, Buffer Size $N$, Scoring Function $F$, Base LLM $\mathcal{M}$
\STATE \textbf{Output:} Final Answer $A_{best}$, Best Reasoning Apth $RP_{best}$
\STATE $B \leftarrow \{\}$  \#\# Initialize empty buffer
\WHILE{$|B| < N$}
    \STATE $(A_i, RP_i) \leftarrow \mathcal{M}(Q)$
    \STATE $SS_i \leftarrow F(A_i, RP_i)$
    \IF{$SS_i \geq T$}
        \STATE $B \leftarrow B \cup \{(A_i, RP_i, SS_i)\}$
    \ENDIF
\ENDWHILE
\STATE $\mathcal{A} \leftarrow \{A | (A, RP, SS) \in B\}$ 
\STATE $A_{best} \leftarrow \underset{A \in \mathcal{A}}{\arg\max} \sum_{(A', RP, SS) \in B: A' = A} SS$
\STATE $RP_{best} \leftarrow \underset{RP}{\arg\max} \{SS | (A_{best}, RP, SS) \in B\}$
\STATE \textbf{return} $A_{best}, RP_{best}$
\end{algorithmic}
\end{algorithm}

\section{Experiments}

\begin{table*}[t]
\centering
\caption{Performance of Different Reasoning Methods on Mathematical, Commonsense, and Symbolic Reasoning Benchmarks, including Chain-of-Thought (CoT), Self-Consistency (SC), Early-Stopping Self-Consistency (ESC), Adaptive Consistency (AC), and Reasoning-Aware Self-Consistency (RASC).  Metrics include Number of Samples Generated ( \# of Gen), Accuracy (Acc), and Gained Accuracy per Sample relative to CoT.}
\label{step_main}
\resizebox{\textwidth}{!}{
\begin{tabular}{c|c|ccc|ccc|ccc}
\toprule
\multirow{2}{*}{Model} & \multirow{2}{*}{Method} & \multicolumn{9}{c}{Benchmark Category} \\
\cmidrule{3-11}
& & \multicolumn{3}{c|}{Mathematical Reasoning} & \multicolumn{3}{c|}{Commonsense Reasoning} & \multicolumn{3}{c}{Symbolic Reasoning} \\
\cmidrule{3-11}
& & \# of Gen & Acc & Gain/Sample & \# of Gen & Acc & Gain/Sample & \# of Gen & Acc & Gain/Sample \\
\midrule
\multirow{5}{*}{GPT-4}
& CoT & 1 & 82.1\% & - & 1 & 83.2\% & - & 1 & 92.5\% & - \\
& SC & 40 & 87.5\% & 0.139\% & 40 & 88.0\% & 0.123\% & 40 & 97.3\% & 0.123\% \\
& ESC & 6.86 (\textcolor{green!50!black}{-82.9\%}) & 87.3\% & 0.889\% & 7.93 (\textcolor{green!50!black}{-80.2\%}) & 88.5\% & 0.767\% & 5.54 (\textcolor{green!50!black}{-86.2\%}) & 97.3\% & 1.056\% \\
& AC & 5.89 (\textcolor{green!50!black}{-85.3\%}) & 87.3\% & 1.065\% & 6.15 (\textcolor{green!50!black}{-84.6\%}) & 87.2\% & 0.779\% & 4.43 (\textcolor{green!50!black}{-88.9\%}) & 97.3\% & 1.395\% \\
& \textbf{RASC} & \textbf{4.59} (\textcolor{green!50!black}{-88.5\%}) & 87.5\% & \textbf{1.503\%} & \textbf{4.74} (\textcolor{green!50!black}{-88.1\%}) & 88.3\% & \textbf{1.367\%} & \textbf{4.19} (\textcolor{green!50!black}{-89.5\%}) & 97.3\% & \textbf{1.500\%} \\
\midrule
\multirow{5}{*}{GPT-3.5 Turbo}
& CoT & 1 & 63.5\% & - & 1 & 71.2\% & - & 1 & 80.1\% & - \\
& SC & 40 & 69.1\% & 0.144\% & 40 & 76.1\% & 0.126\% & 40 & 85.3\% & 0.134\% \\
& ESC & 14.94 (\textcolor{green!50!black}{-62.7\%}) & 69.3\% & 0.417\% & 10.86 (\textcolor{green!50!black}{-72.9\%}) & 76.4\% & 0.527\% & 7.55 (\textcolor{green!50!black}{-81.1\%}) & 86.0\% & 0.903\% \\
& AC & 12.49 (\textcolor{green!50!black}{-68.8\%}) & 67.6\% & 0.357\% & 8.34 (\textcolor{green!50!black}{-79.2\%}) & 75.0\% & 0.517\% & 6.39 (\textcolor{green!50!black}{-84.0\%}) & 84.7\% & 0.853\% \\
& \textbf{RASC} & \textbf{6.62} (\textcolor{green!50!black}{-83.5\%}) & 69.4\% & \textbf{1.054\%} & \textbf{4.95} (\textcolor{green!50!black}{-87.6\%}) & 76.1\% & \textbf{1.238\%} & \textbf{4.36} (\textcolor{green!50!black}{-89.1\%}) & 85.3\% & \textbf{1.547\%} \\
\midrule
\multirow{5}{*}{Vicuna-13B}
& CoT & 1 & 36.2\% & - & 1 & 50.1\% & - & 1 & 41.5\% & - \\
& SC & 40 & 41.7\% & 0.141\% & 40 & 55.9\% & 0.149\% & 40 & 47.3\% & 0.149\% \\
& ESC & 20.29 (\textcolor{green!50!black}{-49.3\%}) & 40.5\% & 0.223\% & 20.89 (\textcolor{green!50!black}{-47.8\%}) & 50.5\% & 0.020\% & 23.05 (\textcolor{green!50!black}{-42.4\%}) & 44.0\% & 0.113\% \\
& AC & 17.44 (\textcolor{green!50!black}{-56.4\%}) & 42.3\% & 0.371\% & 19.92 (\textcolor{green!50!black}{-50.2\%}) & 55.9\% & 0.306\% & 25.77 (\textcolor{green!50!black}{-35.6\%}) & 49.3\% & 0.314\% \\
& \textbf{RASC} & \textbf{8.24} (\textcolor{green!50!black}{-79.4\%}) & 42.0\% & \textbf{0.801\%} & \textbf{7.83} (\textcolor{green!50!black}{-80.4\%}) & 55.0\% & \textbf{0.718\%} & \textbf{8.71} (\textcolor{green!50!black}{-78.2\%}) & 45.3\% & \textbf{0.494\%} \\
\midrule
\multirow{5}{*}{Llama2-7B}
& CoT & 1 & 18.5\% & - & 1 & 63.2\% & - & 1 & 19.5\% & - \\
& SC & 40 & 23.0\% & 0.115\% & 40 & 68.9\% & 0.146\% & 40 & 24.2\% & 0.121\% \\
& ESC & 27.25 (\textcolor{green!50!black}{-31.9\%}) & 23.5\% & 0.190\% & 8.85 (\textcolor{green!50!black}{-77.9\%}) & 68.9\% & 0.726\% & 31.03 (\textcolor{green!50!black}{-22.4\%}) & 24.8\% & 0.177\% \\
& AC & 23.91 (\textcolor{green!50!black}{-40.2\%}) & 21.5\% & 0.131\% & 7.97 (\textcolor{green!50!black}{-80.1\%}) & 68.3\% & 0.735\% & 24.62 (\textcolor{green!50!black}{-38.5\%}) & 24.2\% & 0.199\% \\
& \textbf{RASC} & \textbf{10.71} (\textcolor{green!50!black}{-73.2\%}) & 23.2\% & \textbf{0.484\%} & \textbf{5.11} (\textcolor{green!50!black}{-87.2\%}) & 68.9\% & \textbf{1.390\%} & \textbf{13.09} (\textcolor{green!50!black}{-67.3\%}) & 24.8\% & \textbf{0.439\%} \\
\bottomrule
\end{tabular}
}
\end{table*}

\subsection{Experimetnal Setup}

\noindent \textbf{Baseline and Data:} We compare \texttt{RASC} against three established baseline methods: \texttt{SC} \cite{wang2022self}, \texttt{AC} \cite{aggarwal2023let}, and \texttt{ESC} \cite{li2024escape}. Following the previous work, our main evaluation dataset comprises 16,725 samples spanning commonsense reasoning (Date Understanding, CommonsenseQA), symbolic reasoning (boolean expressions, disambiguation, last letters), and mathematical reasoning (GSM8K, ASDIV, MathQA, SVAMP). To assess generalizability on unseen data, we use out-of-distribution datasets including BigBench \cite{srivastava2023beyond} and StrategyQA \cite{geva2021did}, comprising an additional 10,824 examples that require implicit reasoning steps across various topics. See the Appendix for the configurations of the baseline methods (\ref{app:alg}) and the data (\ref{app:data}).

\begin{figure*}[ht]
\centering
\includegraphics[width=1.0\textwidth]{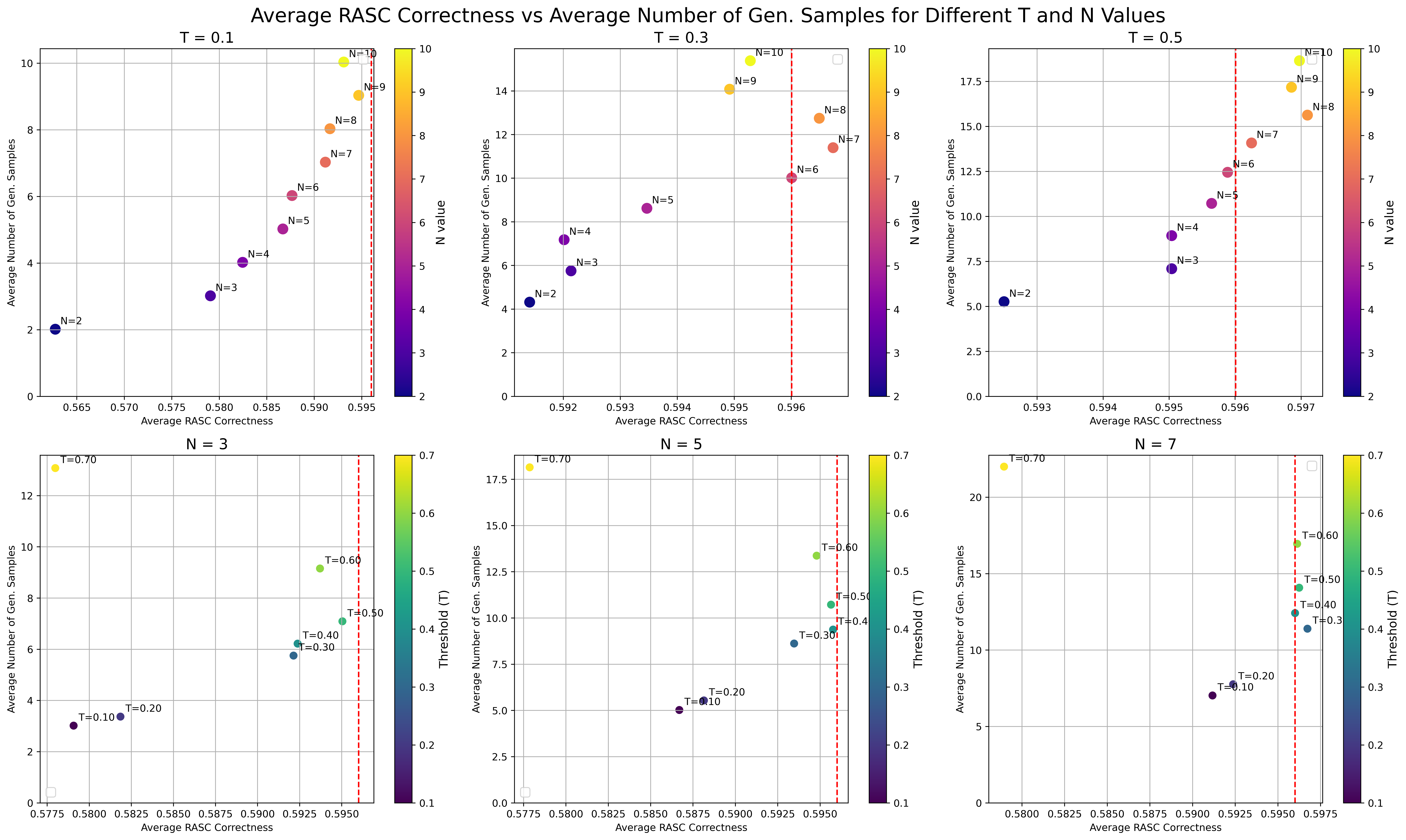}
\caption{Effects of varying $N$ and $T$ on performance tradeoffs, illustrating how changes in these parameters impact both accuracy (RASC Correctness) and the average number of samples generated.}
\label{fig:TN}
\end{figure*}

\noindent \textbf{Implementation Details:} Our experiments utilize \texttt{LLAMA2-7B} \cite{llama3}, \texttt{GPT3.5-turbo/GPT4} \cite{openai2021gpt35turbo}, and \texttt{Vicuna-13B} \cite{vicuna2023}. For parameters of our algorithm, we make buffer size N being \texttt{5} with a quality threshold T of \texttt{0.5} based on experimental results in Figure \ref{fig:TN}. We use customized metrics that measure the trade-off between efficiency and accuracy to select the best hyper-parameters. We use \texttt{Logistic Regression} as a lightweight sufficiency scoring model, which shows a comparable correlation with human judgments and efficiency on running time. We keep the parameters learned from training set and apply it on testset to prevent overfitting. LLM inference is applied with a temperature of \texttt{0.5}. See Appendix \ref{app:exp_details} for further details on experimental setup, including computational resources, model configurations, and prompts. Additional results, including analysis of various base models and Pearson correlation of other NLG metrics with human juedgement, can be found in Appendix \ref{app:add_results}.

\subsection{Main Results}

\textbf{RASC Excels Across Diverse Reasoning Tasks and LLMs: } Table \ref{step_main} showcases RASC's performance across a wide spectrum of reasoning tasks and base Language Models (LLMs). Our comprehensive analysis reveals that RASC consistently achieves superior sample efficiency while maintaining similar accuracy across mathematical, commonsense, and symbolic reasoning tasks. This superior performance and efficiency are statistically significant across multiple benchmarks and models, as detailed in Appendix \ref{subsec:statistical_significance}. The robustness of performance of RASC is further highlighted by its consistent effectiveness across diverse LLMs, from closed-source models like GPT-3.5 Turbo and GPT-4 to open-source alternatives such as Vicuna-13B and Llama2-7B.

\begin{table}[htbp]
\centering
\footnotesize
\caption{Performance and Time Analysis of Methods Using GPT-4. Analysis of other models and costs can be found in Appendix \ref{subsec:additional_performance_analysis}.}
\label{tab:performance_analysis}
\begin{tabularx}{\linewidth}{@{}l *{4}{>{\centering\arraybackslash}X}@{}}
\toprule
\textbf{Method} & \textbf{Accuracy (\%)} & \textbf{Inference Time (s)} & \textbf{Non-Inference Time (s)} & \textbf{Total Time (s)} \\
\midrule
SC   & 90.9 & 398.9 & 0.00 & 398.9 \\
ESC  & 91.0 & 67.8  & 0.04 & 67.9  \\
AC  & 90.6 & 54.9  & 0.06 & 55.0  \\
\textbf{RASC} & \textbf{91.0} & \textbf{45.1} & 2.05 & \textbf{47.2} \\
\bottomrule
\end{tabularx}
\caption*{\small Note: Values are averages per question over all data.}
\vspace{-10pt}
\end{table}

\begin{table*}[htbp]
\centering
\caption{Performance comparison of different feature sets across reasoning tasks. Answer-level features include final answer characteristics. Reasoning-level features capture the quality of the reasoning process. Combined Features incorporate both answer-level and reasoning-level features. See Table \ref{tab:features} for more feature details.}
\label{tab:feature_comparison}
\small
\begin{tabularx}{\textwidth}{@{}l|XX|XX|XX@{}}
\hline
\multirow{2}{*}{Feature Set} & \multicolumn{2}{c|}{Mathematical Reasoning} & \multicolumn{2}{c|}{Symbolic Reasoning} & \multicolumn{2}{c@{}}{CommonSense Reasoning} \\
 & \# Gen & Acc (\%) & \# Gen & Acc (\%) & \# Gen & Acc (\%) \\
\hline
Answer-level only &  10.86 & 55.5 & 7.75 & 58.0 & 6.65 & 74.9 \\
Reasoning-level only & 5.50 & 50.4 & 4.30 & 55.9 & 3.08 & 72.4 \\
Combined Features & \textbf{8.20} & \textbf{55.8} & \textbf{6.28} & \textbf{58.1} & \textbf{5.63} & \textbf{75.0} \\
\hline
\end{tabularx}
\vspace{1ex}
\raggedright
\footnotesize
Results are averaged across all datasets and models. \# Gen: Average number of generated samples. Acc: Accuracy.
\end{table*}

\noindent \textbf{RASC Excels in Computational Efficiency: } Table \ref{tab:performance_analysis}
 empirically highlights RASC's improvement in computational efficiency over other methods. While SC incurs minimum additional computational overhead, it requires the longest inference time. AC and ESC, on the other hand, introduce relatively small overhead due to their stopping criteria calculations. RASC, although introducing the highest non-inference processing time for feature extraction, substantially reduces the overall inference time, which typically constitutes the largest portion of the total computational cost. Notably, RASC reduces inference time by approximately 80\% compared to SC, while maintaining similar accuracy, and outperforming other efficient methods. We thus claim that as long as the non-inference overhead remains relatively smaller than the savings in inference time, RASC will consistently prove more computationally efficient. These gains hold across other evaluated models and is especially significant when inference speed is slower, as demonstrated in Table \ref{tab:performance_analysis_llama2} in the Appendix.

\subsection{Analysis}
\label{AS}
\noindent \textbf{Robust Performance and Tunable Efficiency:} Fig.~\ref{fig:TN} shows RASC's performance across various threshold ($T$) and buffer size ($N$) configurations. RASC demonstrates a unique combination of performance stability and customizable efficiency. Accuracy improves consistently as $T$ increases from 0.1 to 0.5, with the most significant gains between $T=0.1$ and $T=0.3$. Increasing $N$ from 3 to 7 generally enhances accuracy, albeit with diminishing returns. This stability across hyperparameter values showcases RASC's robustness. RASC allows preference-based tuning: lower $T$ and $N$ values prioritize efficiency with minimal accuracy loss, ideal for resource-constrained scenarios, while higher values maximize accuracy at increased computational cost. This dual characteristic of robustness and adaptability enables RASC to optimize for specific accuracy or efficiency requirements across various computational environments.

\noindent \textbf{Performance Analysis on Tasks with Varying Difficulty: }
To further validate RASC's effectiveness across different difficulty levels, we conducted experiments using the MMLU benchmark, examining performance on mathematical topics of varying complexity: Elementary Mathematics, College Mathematics, and Abstract Algebra.

\begin{table}[t]
\centering
\caption{Performance on MMLU Math Categories (GPT-4)}
\label{tab:difficulty_analysis}
\resizebox{\columnwidth}{!}{
\begin{tabular}{l|ccc|ccc|ccc}
\toprule
\multirow{3}{*}{Method} & \multicolumn{3}{c|}{Elementary} & \multicolumn{3}{c|}{College} & \multicolumn{3}{c}{Abstract} \\
& \multicolumn{3}{c|}{Mathematics} & \multicolumn{3}{c|}{Mathematics} & \multicolumn{3}{c}{Algebra} \\
\cmidrule{2-10}
& \#Gen & Acc & Gain/Sample & \#Gen & Acc & Gain/Sample & \#Gen & Acc & Gain/Sample \\
\midrule
CoT & 1 & 95.8 & -- & 1 & 75.4 & -- & 1 & 71.0 & -- \\
SC & 40 & 97.8 & 0.051 & 40 & 84.0 & 0.231 & 40 & 76.0 & 0.128 \\
ESC & \cellcolor{gray!10}4.23 & 97.6 & 0.558 & \cellcolor{gray!10}8.93 & 83.8 & 1.114 & \cellcolor{gray!10}12.45 & 75.8 & 0.442 \\
& {\footnotesize(-89.4\%)} & & & {\footnotesize(-77.7\%)} & & & {\footnotesize(-68.9\%)} & & \\
ASC & \cellcolor{gray!10}3.85 & 97.7 & 0.669 & \cellcolor{gray!10}7.56 & 83.7 & 1.323 & \cellcolor{gray!10}10.28 & 75.7 & 0.510 \\
& {\footnotesize(-90.4\%)} & & & {\footnotesize(-81.1\%)} & & & {\footnotesize(-74.3\%)} & & \\
\rowcolor{gray!20}
\textbf{RASC} & \textbf{3.12} & 97.7 & \textbf{0.904} & \textbf{6.15} & 83.9 & \textbf{1.726} & \textbf{8.54} & 75.9 & \textbf{0.651} \\
& {\footnotesize(-92.2\%)} & & & {\footnotesize(-84.6\%)} & & & {\footnotesize(-78.7\%)} & & \\
\bottomrule
\end{tabular}
}
\caption*{Note: Categories arranged in order of increasing difficulty (Elementary → College → Abstract)}
\end{table}

The results reveal two key patterns. For harder tasks (Abstract Algebra), RASC requires more samples (8.54 vs 6.15 for College Math) to achieve confident stopping, reflecting the reasoning variations in evaluating complex reasoning and result in better performance gains. However, for easier tasks (Elementary Mathematics), RASC achieves extreme sampling efficiency due to rapid recognition of correct reasoning patterns although with a slight less performance improvement. Notably, while harder tasks require more samples, they still maintain significantly reduced sampling compared to SC , demonstrating RASC's ability to balance sampling efficiency with task complexity.

\noindent \textbf{Individual Impact of Reasoning-Level and Answer-Level Features:}  Our comprehensive ablation studies reveal the crucial interplay between reasoning-level and answer-level features across various reasoning tasks. Table~\ref{tab:feature_comparison} demonstrates that \textbf{\emph{integrating both feature types consistently outperforms models using either feature set alone regarding the trade-off between accuracy and number of samples generated, validating our hypothesis of their complementary nature.}} While answer-level features provide a strong baseline, as established in previous work, our novel incorporation of reasoning-level features, providing additional insights form the intermediate reasoning, significantly enhances the model's capabilities in distinguishing better samples, thus reducing the number of generations needed. This synergistic combination enables a more holistic capture of the reasoning process, highlighting the importance of considering both the reasoning path and the final answer in evaluating and optimizing language model outputs.

\noindent \textbf{Selection of High-Fidelity Chains of Thought:} Table \ref{tab:faithfulness} demonstrates RASC's superiority over Self-Consistency (SC) in rationale selection across traditional metrics and manual evaluation, with a 0.7-point increase in Human-Eval Score on a 5-point scale. Qualitative analysis in Appendix also supports this: for a travel distance question (Table \ref{tab:cot_comparison}, Sample 1), RASC provides a clear, correct explanation (human scores: 5) while SC introduces ambiguity (scores: 2). This analysis shows \textbf{\emph{RASC's superior capability in selecting faithful and high-quality Chains of Thought (CoTs)}}, consistently producing more accurate, coherent, and relevant explanations for improved problem-solving.

\begin{table}[ht]
\centering
\caption{Comparison of RASC and SC on CoT Faithfulness. This evaluation used 200 randomly selected questions from our training set. For RASC, we picked the CoT with the highest sufficiency score, and for SC, we selected a random CoT sample that produced the correct final answer. The evaluation assessed BARTScore, CTC, BLURT, and human-evaluated faithfulness scores. For BARTScore, CTC, and BLURT, we manually created golden CoTs as reference.}
\label{tab:faithfulness}
\resizebox{\columnwidth}{!}{%
\begin{tabular}{lccc}
\toprule
\textbf{Metric} & \textbf{RASC} & \textbf{SC} & \textbf{Diff.} \\
\midrule
BARTScore \cite{NEURIPS2021_e4d2b6e6} & 0.61 & 0.39 & +0.22 \\
CTC \cite{deng-etal-2021-compression} & 0.55 & 0.45 & +0.10 \\
BLURT \cite{sellam2020bleurt} & 0.58 & 0.42 & +0.16 \\
Human-Eval Score$^a$ & 4.7 & 4.0 & +0.70 \\
\bottomrule
\end{tabular}%
}
\caption*{\footnotesize $^a$On a scale of 1-5. All other scores range 0-1, representing ranked-based scores where 1 indicates better metrics for RASC over SC and 0 means the other way around. See Appendix \ref{app:human_eval} for details on human evaluation criteria and golden CoT creation process.}
\vspace{-10pt}
\end{table}

 \subsection{Robustness and Generalizability}

\begin{table}[ht]
\centering
\small
\caption{Impact of Different Prompting Strategies on Performance on 100 Random Question from GSM8K dataset, a subset data from Mathematical Reasoning, with GPT-3.5 Turbo (a/b: a = accuracy (\%), b = avg. samples)}
\label{tab:prompt_impact}
\begin{tabular}{@{}lccc@{}}
\toprule
Method & Zero-shot & Few-shot & Least-to-Most \\
\midrule
SC   & 69.0 / 40.0 & 75.0 / 40.0 & 85.0 / 40.0 \\
ESC  & 67.0 / 9.3 & 75.0 / 9.5 & 85.0 / 8.8 \\
AC  & 69.0 / 8.8 & 74.0 / 8.4 & 85.0 / 7.0 \\
\textbf{RASC} & \textbf{69.0} / \textbf{5.1} & \textbf{75.0} / \textbf{5.3} & 85.0 / \textbf{5.0} \\
\bottomrule
\end{tabular}
\end{table}
\noindent \textbf{RASC is consistent on Different Prompts:} To check RASC's robustness to different input prompts, we evaluated RASC's performance using three distinct CoT prompting strategies: zero-shot \cite{kojima2022large}, few-shot \cite{wei2022chain}, and least-to-most \cite{zhou2023leasttomost}.  As Table.\ref{tab:prompt_impact} shows, RASC consistently outperforms other methods across all prompting strategies, demonstrating higher accuracy and lower sample (5 steps on average) requirements. This consistency highlights RASC's adaptability to various ways of eliciting reasoning from language models.

\noindent \textbf{Cross-Domain Adaptability and Performance Scaling:}
To evaluate RASC's generalizability beyond our training distribution for reasoning evaluations, we tested it on two external datasets, including BigBench \cite{srivastava2023beyond} and StrategyQA \cite{geva2021did} comprising 10,824 examples requiring implicit reasoning steps across various topics. Using GPT3.5 turbo as the base model, RASC demonstrated robust performance on these unseen, complex datasets. As illustrated in Figure \ref{fig:ood} in Appendix, \textbf{\emph{RASC achieved the highest accuracy while significantly reducing the average number of samples generated compared to other methods}}. Notably, RASC outperformed SC in accuracy (0.581 vs 0.579) while using only 14.5\% of the samples (5.8 vs 40.0). It also surpassed both ESC and AC in both accuracy and efficiency. This performance suggests RASC's resilience across varied data distributions, highlighting its ability to generalize effectively to novel reasoning tasks without compromising accuracy or computational efficiency.

\section{Conclusion}

Through our investigation of sampling-based LLM reasoning, we discovered that the quality of intermediate reasoning paths serves as a crucial early indicator for efficient sampling decisions. Building on this insight, we developed RASC, a framework that strategically incorporates reasoning path evaluation into the sampling process. Our approach not only achieved a 60-80\% reduction in required samples while maintaining accuracy. By introducing principled methods for assessing reasoning quality and selecting the most faithful rationale from generated samples, RASC provides a systematic solution to the challenge of balancing computational efficiency with reasoning reliability in LLMs. The framework's success across diverse reasoning tasks demonstrates that focusing on the quality of intermediate reasoning, rather than just final answers, offers a promising direction for developing more efficient and trustworthy LLM reasoning systems.

\section{Limitations}
While RASC demonstrates significant improvements in efficiency and faithfulness across various reasoning tasks, it is important to acknowledge some limitations and areas for future research:

\begin{itemize}
    \item \textbf{Feature  Limitations:} While RASC significantly reduces LLM calls, its reliance on manually designed features introduces computational overhead and potential limitations. The feature extraction process adds processing time that, although generally outweighed by savings from reduced LLM calls, could impact performance in scenarios requiring rapid responses or when handling large batches of simple queries. Moreover, these hand-crafted features, while empirically effective, may not capture all aspects of high-quality reasoning across different domains. Future work could explore adaptive strategies that dynamically adjust feature computation based on runtime constraints, investigate lightweight feature approximations, and leverage unsupervised or transfer learning approaches to automatically identify more comprehensive reasoning quality features.
    
    \item \textbf{Faithfulness Evaluation:} While we have shown improvements in faithfulness of the selected chain of thought reasoning, the evaluation relies on automated metrics and limited human evaluation. A more extensive human evaluation could provide stronger evidence on our work.
\end{itemize}

\bibliography{custom}

\appendix

\label{sec:appendix}

\section{Appendix: Experimental Details}
\label{app:exp_details}

\subsection{Base Models}
The language models used in our experiments have varying numbers of parameters:
\begin{itemize}
\item \texttt{LLAMA2-7B}: This model has 8 billion parameters \cite{llama3}.
\item \texttt{GPT3.5-turbo}: The exact number of parameters for this model is not publicly disclosed by OpenAI \cite{openai2021gpt35turbo}.
\item \texttt{Claude-3-Haiku}: The number of parameters for this model is not publicly available from Anthropic \cite{anthropic2024claude3}.
\item \texttt{Vicuna-13B}: This model has 13 billion parameters \cite{vicuna2023}.
\end{itemize}
\subsection{Computational Resources and Infrastructure}
\label{app:comp_resources_infrastructure}

Our experiments were conducted using various models and computing resources. Below, we detail the computational budget for each model and the specifications of our computing infrastructure.

\textbf{Computational Budget}
\begin{itemize}
    \item \texttt{LLAMA2-7B}: We consumed approximately 100 GPU hours for generating all the collected data, running the model, generating CoT samples, and processing the results.
    \item \texttt{GPT3.5-turbo}: Approximately \$500 was spent on API calls to OpenAI, covering all experiments including data generation, CoT samples, and result processing.
    \item \texttt{Claude-3-Haiku}: The total cost for API calls to Anthropic was approximately \$300.
    \item \texttt{Vicuna-13B}: We utilized approximately 150 GPU hours for experiments, including experimental running, inference on CoT samples, and data processing.
\end{itemize}

\textbf{Computing Infrastructure}
Our experiments were conducted on a computing infrastructure equipped with the following hardware:
\begin{itemize}
    \item GPU: NVIDIA GeForce RTX 3070 Ti
    \item CPU: 16 cores 11th Generation Intel Core i7 Processors
    \item RAM: 64GB DDR4
    \item Storage: 1TB NVMe SSD
\end{itemize}

This infrastructure provided the necessary computational power and storage capacity to efficiently run our experiments across various models and datasets.
\subsection{Model Configurations}
\label{app:model_config}
\begin{itemize}
\item \textbf{Temperature}: For all LLM calls, we used a temperature setting of 0.7 to balance creativity and coherence in the generated outputs.
\item \textbf{Max Tokens}: The maximum number of tokens for each response was set to 1024 for consistency across models.
\item \textbf{Top-p (nucleus sampling)}: We used a top-p value of 0.95 for all model calls to ensure diverse yet relevant outputs.
\end{itemize}
\subsection{Algorithm Configurations}
\label{app:alg}
\begin{itemize}
\item \textbf{ESC Algorithm}: We used a window size of 5 for the Entropy-based Stopping Criteria (ESC) algorithm.
\item \textbf{Adaptive Consistency Algorithm}: We used the default settings with stopping\_criteria=BetaStoppingCriteria(0.95) and max\_gens = 40.
\end{itemize}
\subsection{Used Packages}
In this study, several online available Python packages are used to conduct experiments and analysis:
\begin{itemize}
\item NLTK: For calculating Jaccard Similarity, Ngram, tokenizer
\item statistics: For computing logistic regression
\item PyTorch: For using pre-trained LLM
\item pandas: For data manipulation
\item json: Loading and saving JSON data
\item sklearn: For supervised learning models training and evaluation
\item adaptive\_consistency: For implementing adaptive consistency algorithm
\item Levenshtein: For computing Levenshtein distance
\item transformers: For Hugging Face pre-trained model usage
\item LangChain: For LLM API usage and answer parser
\end{itemize}

\subsection{Datasets}
\label{app:data}
To assess the generalizability and robustness of our proposed Reasoning-Aware Self-Consistency (RASC) method, we utilize a diverse range of datasets across multiple reasoning categories. Our evaluation leverages Question Answering datasets spanning three main categories:

\textbf{Mathematical Reasoning:} GSM8K, ASDIV, MathQA, SVAMP

\textbf{Commonsense Reasoning:} Date Understanding, CommonsenseQA

\textbf{Symbolic Reasoning:} Boolean Expressions, Disambiguation, Last Letters

These datasets provide a comprehensive foundation for evaluating our method across various problem domains and task types.

\textbf{Out-of-Domain (OOD) Data:} To rigorously test the adaptability and robustness of our RASC method, we incorporate out-of-domain (OOD) data, specifically StrategyQA \citep{geva2021did} and BigBench \citep{srivastava2023beyond}.
StrategyQA is a question answering benchmark designed for multi-hop reasoning, featuring:
\begin{itemize}
\item 2,780 examples requiring implicit reasoning steps.
\item Diverse, short questions covering a wide range of strategies.
\item Annotations including reasoning step decomposition and supporting evidence.
\item A novel data collection procedure to ensure challenging and diverse questions.
\end{itemize}
BigBench is a comprehensive benchmark consisting of:
\begin{itemize}
\item 204 tasks contributed by 450 authors across 132 institutions.
\item Diverse topics including linguistics, math, common-sense reasoning, and social bias.
\item Tasks designed to be beyond the capabilities of current language models.
\item Evaluations across various model scales and architectures.
\end{itemize}
By using these datasets, we aim to evaluate RASC's performance on questions that require complex, multi-step reasoning strategies not explicitly stated in the question. This allows us to verify the performance of our method on unseen data that demands sophisticated reasoning capabilities, providing insights into its real-world applicability and resilience across different computational environments and language model architectures.
Using GPT3.5 turbo as the base model, RASC demonstrated robust performance on these unseen, complex datasets, comprising a total of 10,824 examples requiring implicit reasoning steps across various topics.

\subsection{Data Processing}

We used a custom data preprocessing pipeline implemented in Python to clean and prepare the input data for our experiments:
\begin{itemize}
\item For text normalization, we employed NLTK's word\_tokenize and WordNetLemmatizer.
\item We used the json\_parser module to parse the results and extract quality features from the model outputs.
\item Our preprocessing pipeline includes steps for handling missing data, removing duplicates, and standardizing text formats.
\end{itemize}
For more detailed information on our data processing techniques, please refer to our GitHub repository \url{https://anonymous.4open.science/r/SC_conf-2D4B/README.md}.

\subsection{Common LLM Errors and Feature Selection}
\label{app:feature}
\subsubsection{Common LLM Errors}
As a crucial part of the solution, a faithful reasoning path is even more important than simply getting the correct answer to the question. Therefore, it is crucial for sampling methods, such as self-consistency, to consider the content quality of the reasoning path itself. To better understand the types of errors that LLM often makes during the response, we utilize common mathematical reasoning and commonsense reasoning datasets to systematically summarize the common errors. We start with the common errors observed by \citet{golovneva2023roscoe} and form an initial set of potential quality features defined in Table~\ref{tab:features-example}. After human evaluation and statistical analysis, the most frequent mistakes that a bad reasoning path has include inconsistency with the question, lack of step coherence (lack of a logical flow), calculation mistakes, and hallucinations. In addition, we observe that LLMs are more likely to hallucinate when the generated context gets significantly longer. In addition, we observe that the RP always leads to an incorrect answer when it 'admits' that a mistake has been made in any of the proposed steps. However, the purpose of this study is not to comprehensively analyzing common mistakes or design automated metrics, we purposely exclude features that require intensive computational cost, such as features that have to be extracted by utilizing external machine learning models (SenteceBert, Transformers, or other neural network-based feature extractors). 
\subsubsection{Answer-level Features} These features include measures such as consistency between consecutive answers ($\text{Local-Consistency}$), consistency with the most common previous answer ($\text{Global-Consistency}$), and detection of deviations from expected response formats ($\text{Parsing-Error}$).

\subsubsection{Reasoning-level Features}

Our novel contribution lies in introducing reasoning-level features, which capture the intrinsic qualities of each Reasoning Path (RP). These features evaluate various aspects of the RP, such as the length and number of reasoning steps ($\text{RP-Length}$, $\text{Num-of-Steps}$), the logical flow between steps ($\text{Step-Relevance}$), and the relevance to the input ($\text{Question-Relevance}$). We also include time depedent features that compare the current RP with previous ones ($\text{Local-Consistency}$, $\text{Global-Consistency}$) and detect self-acknowledged mistakes or uncertainties ($\text{Error-Admitting}$). Refer to Table~\ref{tab:features-example} for more examples. While similar work like ROSCOE \cite{golovneva2023roscoe} introduces sets of reasoning-level features, our approach extends this by incorporating time-dependent features that compare the current RP with previous ones, allowing us to evaluate the consistency and progression of reasoning across multiple steps

\subsection{Preliminary Feature Analysis}

Table \ref{tab:corr_pvalue} presents the correlation coefficients between each feature and the Correctness variable, along with the p-values from t-tests.

\begin{table}[htbp]
\centering
\resizebox{\columnwidth}{!}{%
\begin{tabular}{|l|r|r|}
\hline
\textbf{Feature} & \textbf{Correlation} & \textbf{P-value} \\
\hline
Local-Consistency & 0.367 & 0.00 \\
Global-Consistency & 0.403 & 0.00 \\
RP-Length & -0.0861 & 0.00 \\
Num-of-Steps & -0.0476 & 0.00 \\
Step-Relevance & -0.0264 & 1.25e-169 \\
Question-Relevance & -0.0423 & 0.00 \\
Error-Admitting & -0.0492 & 0.00 \\
Local-Relevance & 0.0530 & 0.00 \\
Global-Relevance & 0.0850 & 0.00 \\
\hline
\end{tabular}%
}
\caption{Correlation with Correctness and P-values (3 significant figures)}
\label{tab:corr_pvalue}
\end{table}

\begin{figure}[htbp]
\centering
\includegraphics[width=0.45\textwidth]{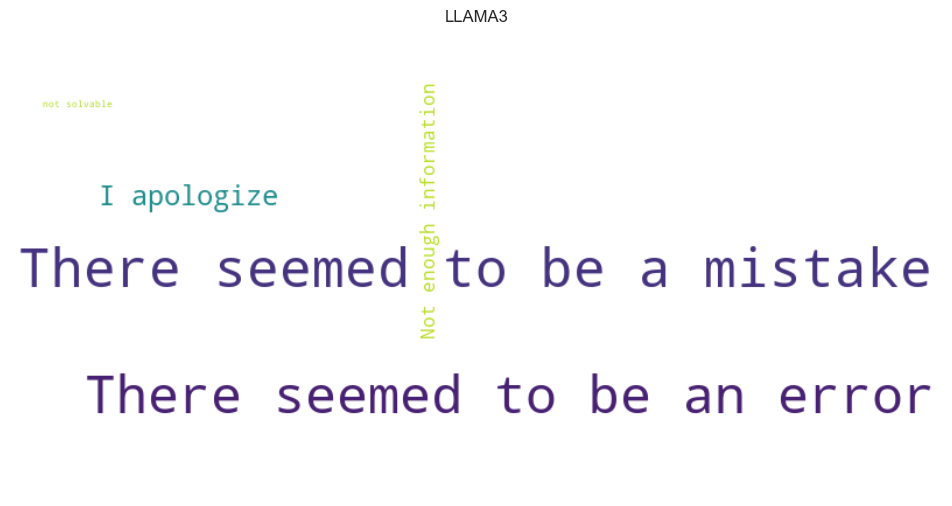}
\caption{Word Cloud of Internal Mistakes for LLAMA3}
\label{fig:llama3_mistakes}
\end{figure}

\begin{figure}[htbp]
\centering
\includegraphics[width=0.45\textwidth]{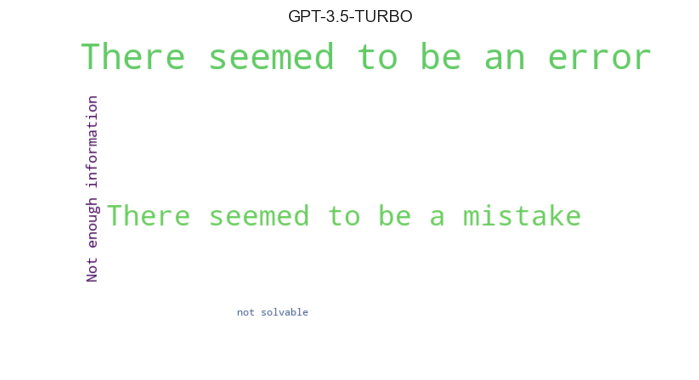}
\caption{Word Cloud of Internal Mistakes for GPT-3.5-TURBO}
\label{fig:gpt3_mistakes}
\end{figure}

Global-Consistency (0.403) and Local-Consistency (0.367) show the strongest positive correlations with Correctness, suggesting these similarity measures between the actual solution and generated content are most predictive of correctness. All features show statistically significant differences between correct and incorrect instances (p-values $\approx$ 0). RP-Length, Error-Admitting, Question-Relevance, Num-of-Steps, and Step-Relevance have weak negative correlations, indicating that longer solutions or those with more steps are slightly less likely to be correct. Word clouds of internal mistakes (Figures \ref{fig:llama3_mistakes} and \ref{fig:gpt3_mistakes}) reveal that both LLAMA3 and GPT-3.5-TURBO models frequently acknowledge potential errors and limitations, using phrases like "There seemed to be a mistake" and "Not enough information". These findings suggest that improving the model's ability to generate solutions similar to the actual solution and encouraging concise responses could enhance performance, while the models' self-awareness of mistakes could be leveraged to improve reliability.

\subsection{Sample Generation Prompts}
\label{sec:prompts}
In this study, we explore three different prompting strategies: Chain of Thought (CoT), Few-Shot, and Least-to-Most. Each strategy aims to enhance the performance of Large Language Models (LLMs) in generating reasoning paths (RPs) for given questions.

\subsubsection{Chain of Thought (CoT) Prompting}
CoT prompting explicitly requires LLMs to reason through the question \textit{step by step}. We utilize zero-shot prompting to generate RPs in our datasets. The prompt is defined as follows:

\begin{tcolorbox}[breakable, enhanced]
\textbf{System Message:} \textit{You are a professional specialized in \{subject\}. You need to help me answer the given question. Notice that you need to solve the question step by step and as detailed as possible. Do not jump to the answer directly. You must follow the RP format instructions.}

\textbf{Human Message:} \textit{\{question\}}
\end{tcolorbox}

\subsubsection{Few-Shot Prompting}
Few-shot prompting provides the LLM with examples of how to solve problems using step-by-step reasoning. This approach helps guide the model in generating structured and detailed responses. The prompt is structured as follows:

\begin{tcolorbox}[breakable, enhanced]
\textbf{System Message:} \textit{You are a professional specialized in \{subject\}. Your task is to answer the given question using step-by-step reasoning. Follow the examples provided, breaking down your thought process into clear steps before providing the final answer.}

\textbf{Human Message:} \textit{Here are two examples of how to solve problems using step-by-step reasoning:}

\textit{[Example 1 and Example 2 are provided here, demonstrating step-by-step problem-solving]}

\textit{Now, please solve the following question using a similar step-by-step approach: \{question\}}
\end{tcolorbox}

\subsubsection{Least-to-Most Prompting}
The Least-to-Most approach involves breaking down complex problems into simpler sub-problems, solving them in order of increasing complexity, and using those solutions to address the main question. This strategy is particularly useful for tackling intricate problems. The prompt is structured as follows:

\begin{tcolorbox}[breakable, enhanced]
\textbf{System Message:} \textit{You are an expert problem solver specialized in \{subject\}. Your task is to break down and solve complex problems using the Least-to-Most approach. This means you'll divide the main problem into simpler sub-problems, solve them in order of increasing complexity, and use those solutions to address the main question.}

\textbf{Human Message:} \textit{Let's solve problems using the Least-to-Most approach. Here's an example:}

\textit{[An example demonstrating the Least-to-Most approach is provided here]}

\textit{Now, please use this Least-to-Most approach to solve the following problem. Break it down into simpler sub-problems, solve them in order, and then use those solutions to address the main question: \{question\}}
\end{tcolorbox}

Each of these prompting strategies aims to elicit detailed, step-by-step reasoning from the LLM, enabling us to generate high-quality reasoning paths for analysis and comparison.

\subsection{Human Evaluation Criteria}
\label{app:human_eval}

\noindent\textbf{Annotator Profile:} The evaluation was conducted by a group of 10 individuals: 7 males and 3 females. The age range of the annotators was 23-30 years old, with a median age of 25. All annotators hold technical degrees spanning mathematics, computer science, and statistics, ensuring their capability to comprehend the logic behind the generated rationales effectively. Their educational backgrounds are as follows:
\begin{itemize}
    \item 1 Assistant Professors (1 in Computer Science)
    \item 6 PhD candidates (1 in Computer Science, 4 in Data Science, 1 in Statistics)
    \item 3 Master's students (1 in Data Science, 1 in Electrical Engineering, 1 in Computational Mathematics)
\end{itemize}

Geographically, the annotators represent two countries:
\begin{itemize}
    \item 3 from the United States
    \item 7 from China
\end{itemize}

Professionally, 8 are currently in academia (1 assistant professors, 5 PhD candidates, 2 Master's students), while 2 work in industry with Master's degrees and 1-3 years of relevant work experience related to technology.

\textbf{Ethical Considerations and Consent:} Prior to participating in the evaluation process, all annotators were provided with information detailed the study's objectives, the nature of their participation, and how their data would be utilized. Specifically, annotators were informed that their ratings, demographic information, and any generated "golden" CoTs would be used for research purposes and potentially published anonymously. They were assured of the confidentiality of their personal information and the anonymization of their responses in any resulting publications. Annotators were also informed of their right to withdraw from the study at any point and request the removal of their data. The data retention policy, including the duration of data storage and the method of eventual deletion, was clearly outlined. Furthermore, annotators were provided with contact information for the research team to address any concerns or questions about their data usage. 

\textbf{Evaluation Instruction:} 
Human evaluators were asked to rate the generated Chain of Thought (CoT) responses on two main criteria:

\begin{enumerate}
    \item \textbf{Overall Quality} [1-5]: This criterion assesses whether the generated response answers the question in a well-justified manner. The scale is interpreted as follows:
    \begin{itemize}
        \item 1: Incomprehensible and completely incorrect. The response is unintelligible or entirely unrelated to the question.
        \item 2: Mostly incorrect with major flaws in reasoning. The response may be partially related to the question but contains significant errors or misunderstandings.
        \item 3: Partially correct but with notable gaps or minor errors. The response addresses the question but lacks full justification or contains some inaccuracies.
        \item 4: Mostly correct with minor imperfections. The response is well-justified and accurate, with only slight room for improvement.
        \item 5: Clear, correct, and fully justified. The response comprehensively answers the question with sound reasoning and no errors.
    \end{itemize}
    
    \item \textbf{Coherency} [1-5]: This criterion evaluates whether the whole generated response makes sense, regardless of its correctness in addressing the context. The scale is interpreted as follows:
    \begin{itemize}
        \item 1: Completely incoherent. The response is a jumble of words or concepts with no logical flow or structure.
        \item 2: Mostly incoherent with occasional comprehensible parts. The response has severe issues with logical flow, making it very difficult to follow.
        \item 3: Partially coherent but with noticeable lapses in logic or structure. The response has a basic structure but contains confusing or disconnected elements.
        \item 4: Mostly coherent with minor clarity issues. The response has a clear logical flow with only slight ambiguities or structural weaknesses.
        \item 5: Perfectly coherent and easy to understand. The response has a clear, logical structure with smooth transitions between ideas.
    \end{itemize}
\end{enumerate}

The overall score for each evaluated response is calculated as the average of these two metrics.

In addition to manually evaluating the best CoT responses (based on our method) and a randomly chosen CoT from SC, evaluators were also tasked with creating a "golden" CoT when necessary so that we have a reference available to calculate the BARTScore \cite{NEURIPS2021_e4d2b6e6}. This process involved:

\begin{enumerate}
    \item Reviewing a random CoT for human-like reasoning.
    \item Assessing whether the flow of logic closely resembles what a human solver would do, considering:
    \begin{itemize}
        \item Natural, step-by-step reasoning processes
        \item Logical order of steps
        \item Inclusion of relevant intermediate calculations or considerations
        \item Appropriate level of detail for human-like problem-solving
    \end{itemize}
    \item Rewriting the CoT if it did not closely resemble human-like reasoning, ensuring:
    \begin{itemize}
        \item Maintenance of the correct answer while improving the reasoning process
        \item Structure reflecting human approach to problem-solving
        \item Clear, concise, and easy-to-follow response
    \end{itemize}
\end{enumerate}

This comprehensive evaluation process ensures a thorough assessment of the quality and coherency of generated CoT responses, as well as the creation of human-like golden standards when needed. See Table \ref{tab:cot_comparison} for two examples.

\begin{table*}[ht]
\centering
\caption{Comparison of Best CoT, SC CoT, and Golden CoT with Human Evaluation}
\label{tab:cot_comparison}
\resizebox{\textwidth}{!}{%
\begin{tabular}{p{0.05\textwidth}p{0.15\textwidth}p{0.05\textwidth}p{0.2\textwidth}p{0.2\textwidth}p{0.2\textwidth}cc}
\toprule
Sample & Question & Answer & RASC Rationale & SC Rationale & Golden CoT & \multicolumn{2}{c}{Ratings (Qua, Coh)} \\
\cmidrule(lr){7-8}
 &  &  &  &  &  & Best & SC \\
\midrule
1 & Elvis starts driving from his house and travels west for 5 hours. Then he turns around and travels east for 8 hours. If he was driving at an average speed of 18mph for both parts of the journey, how far is he from his house now? & 54 miles & If Elvis traveled west for 5 hours at 18mph, he covered 5 x 18 = 90 miles. When he traveled east for 8 hours at 18mph, he covered 8 x 18 = 144 miles. The distance he is from his house now is the difference between the two distances, which is 144 - 90 = 54 miles. The answer is 54 miles. & If Elvis drove west for 5 hours at 18mph, he covered a distance of 5 * 18 = 90 miles. When he turned around and drove east for 8 hours at 18mph, he covered a distance of 8 * 18 = 144 miles. The total distance he covered is 90 + 144 = 234 miles. However, we do not know the direction of his house. If he is now east of his house, he is 144 - 90 = 54 miles away. If he is now west, he is 90 - 144 = -54 miles away. So, we need more information to determine the distance he is from his house. & Elvis drove for 5 hours at 18mph, so he covered 5 * 18 = 90 miles to the west. He then turns around and drives for 8 hours at 18mph, so he covers 8 * 18 = 144 miles to the east. The total distance from his house is 144 - 90 = 54 miles to the east. The answer is 54 miles. & 5, 5 & 1, 3 \\
\midrule
2 & Are the founders of Skype from Asia? & No & The founders of Skype are from Europe, not Asia. The founders are Estonians Ahti Heinla, Priit Kasesalu, and Jaan Tallinn. So the answer is no. & The founders of Skype are Niklas Zennström and Janus Friis, both of whom are from Europe (Sweden and Denmark, respectively). Therefore, the answer is no. & The founders of Skype are from Europe. The founders of Skype are Estonians Ahti Heinla and Priit Kasesalu. Estonia is a country located in Northern Europe, not Asia. So the answer is no. & 4, 5 & 2, 4 \\
\bottomrule
\end{tabular}%
}
\caption*{Note: Qua = Quality, Coh = Coherency. Ratings are on a scale of 1-5, with 5 being the highest.}
\end{table*}

\subsection{Evaluation Metric for Accuracy-Cost Trade-Off}
\label{app:acc_cost_metric}

To quantitatively assess the trade-off between accuracy and cost, we introduced a custom metric that balances both factors. This metric normalizes the accuracy and cost values and computes a weighted average to provide a single score representing the overall performance of a method.

Let $\text{acc}$ denote the accuracy of a method, and $\text{cost}$ denote the computational cost, measured as the number of samples generated. Additionally, let $\text{sc\_acc}$ and $\text{sc\_cost}$ represent the accuracy and cost of the Self-Consistency (SC) method, respectively. Let $\text{single\_sample\_acc}$ and $\text{direct\_cost}$ denote the accuracy from using only the first sample's answer and the cost of the direct sampling method, respectively.

The metric is defined as:

\begin{equation}
\text{metric} = 0.5 \times \text{acc\_factor} + 0.5 \times \text{cost\_factor}
\end{equation}

where $\text{acc\_factor}$ and $\text{cost\_factor}$ are normalized values of accuracy and cost, calculated as follows:

\begin{equation}
\text{acc\_factor} = 
\begin{cases}
1 & \text{if } \text{acc} \geq \text{sc\_acc} \\
0 & \text{if } \text{acc} \leq \text{sam\_acc} \\
\frac{\text{acc} - \text{sam\_acc}}{\text{sc\_acc} - \text{single\_acc}} & \text{otherwise}
\end{cases}
\end{equation}

\begin{equation}
\text{cost\_factor} = 
\begin{cases}
1 & \text{if } \text{cost} \leq \text{direct\_cost} \\
0 & \text{if } \text{cost} \geq \text{sc\_cost} \\
\frac{\text{sc\_cost} - \text{cost}}{\text{sc\_cost} - \text{dir\_cost}} & \text{otherwise}
\end{cases}
\end{equation}

The $\text{acc\_factor}$ is normalized to be between 0 and 1, where 1 corresponds to an accuracy greater than or equal to the SC method, and 0 corresponds to an accuracy less than or equal to the accuracy using just the first sample's answer. Similarly, the $\text{cost\_factor}$ is normalized to be between 0 and 1, where 1 corresponds to a cost less than or equal to the direct sampling method, and 0 corresponds to a cost greater than or equal to the SC method.
The metric function calculates the weighted average of $\text{acc\_factor}$ and $\text{cost\_factor}$, giving equal importance to both accuracy and cost. A higher value of the metric indicates a better trade-off between accuracy and cost. This function enables us to compare different methods and configurations based on their ability to achieve high accuracy while minimizing computational cost.

\subsection{Reproducibility}
To ensure reproducibility of our results:
\begin{itemize}
\item We set a fixed random seed (42) for all randomized processes.
\item All experiments were run using Python 3.8.10 in a conda environment. The full list of dependencies and their versions can be found in the \texttt{requirements.txt} file in our repository.
\end{itemize}

 \section{Appendix: Additional Results}
\label{app:add_results}

\subsection{Impact of Base Estimator:} 
\label{app:base_model}
The performance of different sufficiency scoring models, as summarized in Table \ref{tab:model_comparison}, identifies Logistic Regression as the most effective model in terms of accuracy and sample utilization. This finding shows the importance of selecting an appropriate sufficiency scoring model to maximize the benefits of the RASC approach. These results are obtained by training the sufficiency scoring models on the training set and then fine-tuning the best set of hyperparameters (N and T) using the customized metrics. The final evaluation is performed on the test set to ensure the robustness of the model.

\begin{table}[htbp]
\centering
\caption{Performance Comparison of Sufficiency Scoring Models}
\label{tab:model_comparison}
\resizebox{\columnwidth}{!}{%
\begin{tabular}{@{}lcc@{}}
\toprule
\textbf{Model} & \textbf{Accuracy (\%)} & \textbf{Num. Samples} \\
\midrule
Random & 41.9 & 9.77 \\
NB & 45.9 & 7.91 \\
\textbf{LR} & \textbf{46.0} & \textbf{7.87} \\
RF & 45.8 & 8.38 \\
HHEM\textsuperscript{} & 42.4 & 9.92 \\
\bottomrule
\end{tabular}%
}
\caption{\footnotesize{\textsuperscript{*}HHEM: Hallucination Detector Model based on microsoft/deberta-v3-base, initially trained on NLI data}}
\end{table}

Among the sufficiency scoring models evaluated \textbf{Logistic Regression} achieves the highest accuracy of \texttt{46.0\%} while requiring the lowest average number of samples (\texttt{5.87}) compared to other models. This indicates that it effectively captures the relationship between the features extracted from the RP and the likelihood of hallucination, which allows RASC to make informed decisions during the weighted majority voting process.

The accuracy of Random model drops below the baseline of Self-Consistency. The poor performance of the Random model, which assigns sufficiency scores randomly, emphasizes the importance of using a well-designed scoring model in the RASC approach. Without a meaningful assessment of the generated content's quality and consistency, the weighted majority voting process cannot effectively distinguish between reliable and unreliable samples, leading to suboptimal results.

The HHEM Model, which is a DeBERTa-based hallucination detector sourced from Hugging Face \cite{honovich2022true}, does not perform as well as the other models in this context. This suggests that relying solely on a pre-trained model without considering the specific characteristics and requirements of the RASC approach may not yield the best results. The superior performance of models like Logistic Regression (LR), Naive Bayes (NB), and Random Forest (RF), which utilize manual feature engineering tailored to the RASC framework, highlights the importance of crafting features that capture the nuances of the generated content as we discussed in the method section.

\subsection{Statistical Significance}
\label{subsec:statistical_significance}

To rigorously evaluate the performance of RASC, we conducted statistical significance experiments across multiple benchmarks using various language models. The comparison between RASC (Reasoning-Aware Self-Consistency) and SC (Self-Consistency) methods reveals interesting patterns across different reasoning types. In terms of correctness, Tables \ref{tab:correctness_comparison_rasc_sc} show no statistically significant differences between RASC and SC across common sense, mathematical, and symbolic reasoning tasks. However, the steps comparison in Tables \ref{tab:steps_comparison_rasc_sc} demonstrates that RASC consistently requires significantly fewer steps than SC across all reasoning types, with p-values of 0 indicating strong statistical significance. This holds the same for AC and RASC comparison as demonstrated in Table \ref{tab:correctness_comparison} and Table \ref{tab:steps_comparison} These results suggest that while RASC maintains comparable correctness to SC, it achieves this with substantially improved efficiency in terms of the number of steps required.

\begin{table}[htbp]
\centering
\footnotesize
\caption{Performance and Time Analysis of Methods Using Llama2.}
\label{tab:performance_analysis_llama2}
\begin{tabularx}{\linewidth}{@{}lXXXX@{}}
\toprule
\textbf{Method} & \textbf{Accuracy (\%)} & \textbf{Inference Time (s)} & \textbf{Non-Inference Time (s)} & \textbf{Total Time (s)} \\
\midrule
SC   & 38.7 & 560.00 & 0.01 & 560.01 \\
ESC  & 39.1 & 313.32 & 0.40 & 313.72 \\
AC  & 38.0 & 263.62 & 0.54 & 264.16 \\
\textbf{RASC} & 39.0 & \textbf{134.96} & 2.05 & \textbf{137.01} \\
\bottomrule
\end{tabularx}
\caption*{\small Note: All values are averages per 1 question.}
\vspace{-10pt}
\end{table}

\begin{table}[ht]
\centering
\caption{Pearson Correlation of Different Metrics with Human Ratings}
\label{tab:nlg_metrics}
\begin{tabular}{lc}
\hline
\textbf{Metric} & \textbf{Pearson Correlation} \\
\hline
RASC Score & 0.68 \\
BARTScore & 0.52 \\
BLURT & 0.58 \\
CTC & 0.65 \\
\hline
\end{tabular}
\caption*{\small Note: Higher correlation indicates closer alignment with human judgments. All correlations are statistically significant (p < 0.01).}
\end{table}

\begin{table*}[t]
\centering
\caption{Mean Difference, P-value, 95\% Confidence Interval, and Significance of Correctness Comparison between RACS and AC Methods on Various Reasoning Types}
\label{tab:correctness_comparison}
\begin{tabular}{|l|c|c|c|c|}
\hline
\textbf{Comparison} & \textbf{Mean Difference} & \textbf{P-value} & \textbf{95\% CI} & \textbf{Significant} \\
\hline
RACS vs AC (common sense reasoning) & -0.0075 & 0.6697 & [-0.0565, 0.0414] & No \\
\hline
RACS vs AC (mathematical reasoning) & -0.004 & 0.745 & [-0.0384, 0.0383] & No \\
\hline
RACS vs AC (symbolic reasoning) & 0.0033 & 0.9044 & [-0.0738, 0.0805] & No \\
\hline
\end{tabular}
\end{table*}

\begin{table*}[t]
\centering
\caption{Mean Difference, P-value, 95\% Confidence Interval, and Significance of Steps Comparison between RACS and AC Methods on Various Reasoning Types}
\label{tab:steps_comparison}
\begin{tabular}{|l|c|c|c|c|}
\hline
\textbf{Comparison} & \textbf{Mean Difference} & \textbf{P-value} & \textbf{95\% CI} & \textbf{Significant} \\
\hline
RACS vs AC (common sense reasoning) & -3.4301 & 0 & [-4.2164, -2.6439] & Yes \\
\hline
RACS vs AC (mathematical reasoning) & -7.4156 & 0 & [-8.0905, -6.7408] & Yes \\
\hline
RACS vs AC (symbolic reasoning) & -7.7112 & 0 & [-9.4171, -6.0052] & Yes \\
\hline
\end{tabular}
\end{table*}

\begin{table*}[t]
\centering
\caption{Mean Difference, P-value, 95\% Confidence Interval, and Significance of Correctness Comparison between RASC and SC Methods on Various Reasoning Types}
\label{tab:correctness_comparison_rasc_sc}
\begin{tabular}{c|c|c|c|c}
\toprule
\textbf{Comparison} & \textbf{Mean Difference} & \textbf{P-value} & \textbf{95\% CI} & \textbf{Significant} \\
\midrule
RASC vs SC (common sense reasoning) & -0.0075 & 0.6697 & [-0.0565, 0.0414] & No \\
RASC vs SC (mathematical reasoning) & -0.0068 & 0.5819 & [-0.0412, 0.0275] & No \\
RASC vs SC (symbolic reasoning) & 0.0067 & 0.8104 & [-0.0705, 0.0839] & No \\
\bottomrule
\end{tabular}
\end{table*}

\begin{table*}[t]
\centering
\caption{Mean Difference, P-value, 95\% Confidence Interval, and Significance of Steps Comparison between RASC and SC Methods on Various Reasoning Types}
\label{tab:steps_comparison_rasc_sc}
\begin{tabular}{c|c|c|c|c}
\toprule
\textbf{Comparison} & \textbf{Mean Difference} & \textbf{P-value} & \textbf{95\% CI} & \textbf{Significant} \\
\midrule
RASC vs SC (common sense reasoning) & -34.7975 & 0 & [-35.0068, -34.5882] & Yes \\
RASC vs SC (mathematical reasoning) & -32.5354 & 0 & [-32.7298, -32.3409] & Yes \\
RASC vs SC (symbolic reasoning) & -32.4240 & 0 & [-32.9474, -31.9007] & Yes \\
\bottomrule
\end{tabular}
\end{table*}

\smallskip
\footnotesize{Note: SC stands for Self-Consistency, a baseline method for comparison.}
\normalsize

\subsection{Additional Performance and Time Analysis}
\label{subsec:additional_performance_analysis}

This section focuses on the performance and time analysis of the open-source Llama2 model.

Table \ref{tab:performance_analysis_llama2} compares different methods using the Llama2 model. RASC achieves the highest accuracy (38.86\%) while significantly reducing inference time. Compared to SC, RASC improves accuracy by 0.33 percentage points and reduces total time by 66.23

These results demonstrate RASC's effectiveness in enhancing both performance and time efficiency. While we don't provide direct cost analysis for closed-source models, the substantial reduction in inference time achieved by RASC would likely translate to significant cost savings in scenarios where API calls are charged based on usage time.

\subsection{Comparison with Existing NLG Metrics:} 
\label{subsec:nlg_metrics_comparison}
To evaluate the effectiveness of our sufficiency scoring method, we compared it against three popular automated metrics for natural language generation: BARTScore \citep{NEURIPS2021_e4d2b6e6}, BLURT \citep{sellam2020bleurt}, and CTC \citep{deng-etal-2021-compression}. Using the same 100 samples from the faithfulness analysis, we calculated the Pearson correlation between these scores and human-annotated quality ratings.

As shown in Table \ref{tab:nlg_metrics}, RASC's sufficiency scores demonstrate a higher correlation with human ratings (0.68) compared to BARTScore (0.52), BLURT (0.58), and CTC (0.65). This suggests that our scoring method more closely aligns with human judgments of CoT quality in reasoning tasks, validating its effectiveness in assessing the quality of generated reasoning paths. The substantial improvement over BARTScore and the slight edge over BLURT and CTC underscore RASC's capability to capture nuances in reasoning quality that align more closely with human evaluations.

\subsection{Test Data Results}
\label{app:strategyqa_results}
\begin{figure*}[htbp]
\centering
\includegraphics[width=0.8\textwidth]{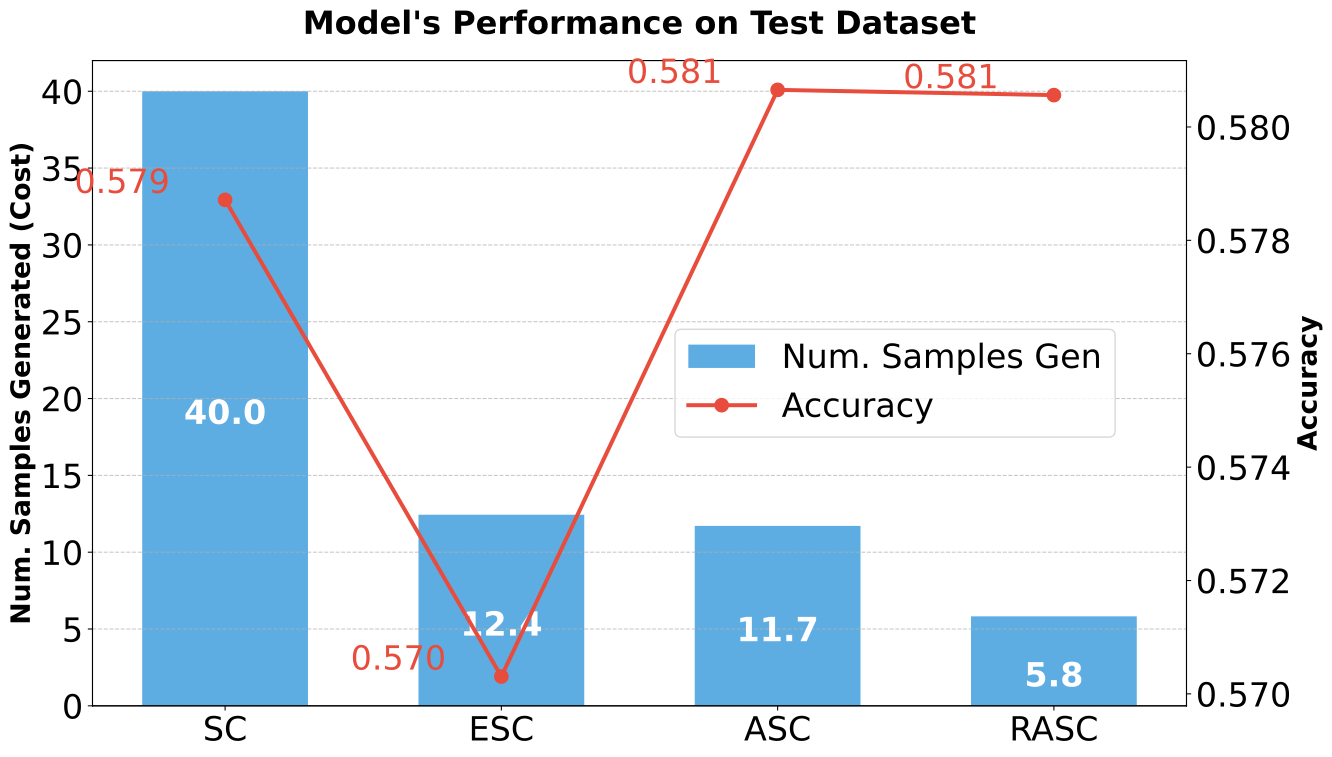}
\caption{Performance of models on out-of-distribution datasets using GPT-3.5 Turbo, assessing method's robustness across diverse reasoning tasks.}
\label{fig:ood}
\end{figure*}

To evaluate RASC's generalizability,  Fig.~\ref{fig:ood} presents the performance comparison between RASC and other methods on this dataset.

\begin{table*}[h!]
    \centering
    \caption{Answer-Level and Reasoning-Level Feature Extraction Example.}
    \label{tab:features-example}
    \small  
    \resizebox{\textwidth}{!}{  
    \begin{tabular}{>{\raggedright\arraybackslash}p{0.22\textwidth}  
                    >{\raggedright\arraybackslash}p{0.70\textwidth}} 
        \toprule
        \textbf{Feature} & \textbf{Example} \\
        \midrule
        \multicolumn{2}{l}{\textbf{Answer-Level Features}} \\  
        \cmidrule(l){1-2} 
        \textbf{Local-Consistency} ($a_{t} \leftrightarrow a_{t-1}$) & Ans1 = 3, Ans2 = 2 $\rightarrow$ Local-Consistency = 0; Ans1 = 3, Ans2 = 3 $\rightarrow$ Local-Consistency = 1 \\
        \textbf{Global-Consistency} ($a_{t} \leftrightarrow a_1,...,a_{t-1}$)&Prev\_ans = [A,B], Cur\_ans = A $\rightarrow$ Global-Consistency = 1; Prev\_ans = [A,B], Cur\_ans = C $\rightarrow$ Global-Consistency = 0 \\
        \textbf{Parsing-Error} ($a_t$) &  Ans: error $\rightarrow$ Parsing-Error = 1; Ans: 2 $\rightarrow$ Parsing-Error = 0 \\
        \midrule
        \multicolumn{2}{l}{\textbf{Reasoning-Level Features}} \\  
        \cmidrule(l){1-2} 
        \textbf{RP-Length} & RP: "I love LLM" $\rightarrow$ RP-Length = 3\\
        \textbf{Num-of-Steps} ($r_t$)  & RP: "Step1:xxx, Step2:xxx" $\rightarrow$ Num-of-Steps = 2 \\
        \textbf{Step-Relevance} ($r_t$) & Step1: Paper reviewer loves LLMs; Step2: They use LLM to review my paper $\rightarrow$ Step-Relevance = 3/7 \\
        \textbf{Question-Relevance} ($r_t \leftrightarrow Q$) & Question: John loves lego, RP: Bob has 2 legos $\rightarrow$ Question-Relevance = 0.33; Question: John loves lego, RP: John has 2 legos $\rightarrow$ Question-Relevance = 0.66 \\
        \textbf{Error-Admitting} ($r_t$) & RP: It seems that I made a mistake in the previous steps $\rightarrow$ Error-Admitting = 1 \\
        \textbf{Local-Relevance} ($r_t \leftrightarrow r_{t-1}$) & RP1: I love LLM, RP2: I love LLM $\rightarrow$ Local-Relevance = 1; RP1: I love LLM, RP2: Bob hates LLM $\rightarrow$ Local-Relevance = 0.33 \\
        \textbf{Global-Relevance} ($r_t \leftrightarrow r_1, ..., r_{t-1}$) & RP1-2: [I love LLM, LLM is good], RP3: I love LLM $\rightarrow$ Global-Relevance = 0.67; RP1-2: [I love LLM, LLM is good], RP3: Life is good $\rightarrow$ Global-Relevance = 0.33 \\
        \bottomrule
    \end{tabular}
    }
\end{table*}

\section{Theoretical Analysis of Reasoning-Aware Self-Consistency}
\label{app:theory_rasc}

In this section, we provide a formal theoretical justification for the design of the Reasoning-Aware Self-Consistency (RASC) framework. Our analysis is grounded in information theory and demonstrates how RASC optimizes inference-time reasoning through a dual information gain framework that incorporates both answer consistency and reasoning path quality.

\subsection{Preliminaries and Notation}

Let $A$ denote the set of possible answers and let $p(a)$ be the probability mass function over $A$. The entropy of the answer distribution is given by:

\[
H(A) = -\sum_{a \in A} p(a) \log p(a).
\]

A higher entropy $H(A)$ implies greater uncertainty in the answer distribution, which corresponds to lower consistency. The goal of self-consistency-based inference methods is to iteratively refine the answer distribution such that $H(A)$ decreases over successive samples.

Now, let $R$ denote the set of all possible reasoning paths. Given an answer $a \in A$, we define the conditional entropy of reasoning paths given an answer as:

\[
H(R|A) = -\sum_{a \in A} p(a) \sum_{r \in R} p(r | a) \log p(r | a).
\]

This quantity measures the uncertainty in the reasoning paths that lead to a particular answer. Since RASC is designed to enhance the coherence and quality of reasoning paths, it aligns with the objective of reducing $H(R|A)$ after an answer has been proposed.

\subsection{Information Gain and Sampling Efficiency}

The information gain from each sample is defined as the reduction in uncertainty of the answer distribution:

\[
IG = H(A_{\text{prior}}) - H(A_{\text{posterior}}).
\]

For classical self-consistency methods (e.g., Self-Consistency (SC), Early Stopping Self-Consistency (ESC), Adaptive Consistency (AC)), the information gain is driven solely by answer consistency. That is, we define:

\[
IG_{\text{ans}} = f(c_i),
\]

where $c_i$ is a consistency metric for the $i$-th sample, and $f: \mathbb{R} \to \mathbb{R}$ is a function characterizing how consistency influences information gain. The specific form of $f$ varies across methods: for instance, ESC applies a sliding window to assess convergence, whereas AC employs an adaptive sequence check.

In contrast, RASC incorporates reasoning path quality into the information gain model. Specifically, the information gain is given by:

\[
IG_{\text{ans-RP}} = f(c_i) + g(r_i),
\]

where $r_i$ represents the quality of the reasoning path associated with the $i$-th sample, and $g: \mathbb{R} \to \mathbb{R}$ is a function capturing the contribution of reasoning coherence and correctness to overall information gain. The augmentation by $g(r_i)$ ensures that the refinement of both answer consistency and reasoning quality contributes to entropy reduction.

\subsection{Stopping Condition and Convergence}

The sampling process in RASC is governed by an optimization criterion where sampling terminates when the cumulative information gain exceeds a predefined threshold $T$:

\[
\sum_{i=1}^{n} IG_i \geq T.
\]

Given the dual contribution of answer consistency and reasoning coherence in RASC, the total entropy $H(A) + H(R|A)$ decreases at a faster rate compared to answer-only self-consistency methods. That is, for a given number of samples $n$, we have:

\[
H_{\text{RASC}}(A, R | n) < H_{\text{SC}}(A | n),
\]

where $H_{\text{RASC}}(A, R | n)$ and $H_{\text{SC}}(A | n)$ denote the total entropy under RASC and SC, respectively, after $n$ samples. This inequality formalizes the efficiency of RASC, demonstrating that fewer samples are required to reach the stopping criterion.

\subsection{Implications and Conclusion}

The theoretical analysis yields the following key conclusions:
\begin{enumerate}
    \item \textbf{Reduction in Entropy:} Answer-level features contribute to minimizing $H(A)$, while reasoning-level features contribute to minimizing $H(R|A)$. The combination ensures a more rapid decrease in overall uncertainty.
    \item \textbf{Enhanced Information Gain:} The presence of $g(r_i)$ in RASC ensures that each sample contributes more effectively to uncertainty reduction, thereby accelerating convergence.
    \item \textbf{Empirical Consistency:} Our results confirm that RASC requires fewer samples than ESC and AC while maintaining accuracy, implying that the expected information gain from reasoning evaluation $g(r_i)$ satisfies:

   \[
   \mathbb{E}[g(r_i)] > 0.
   \]
   \item \textbf{Sufficiency Function Validity}: The sufficiency function used in RASC effectively captures reasoning quality, ensuring that the decision to stop sampling is based on a meaningful reduction in uncertainty.

\end{enumerate}

While the current formulation of $g(r_i)$ has demonstrated empirical success, future research could explore optimizing its structure via feature selection or more sophisticated scoring functions.
\end{document}